\documentclass[english,linesnumbered,commentsnumbered,ruled,vlined]{article}
\usepackage{subfig}
\usepackage[T1]{fontenc}
\usepackage[latin9]{inputenc}
\usepackage{booktabs}
\usepackage{algorithm2e}
\usepackage{amsmath}
\usepackage{amsthm}
\usepackage{amssymb}
\usepackage{graphicx}
\usepackage[authoryear]{natbib}

\makeatletter

\providecommand{\tabularnewline}{\\}

\theoremstyle{plain}
\newtheorem{thm}{\protect\theoremname}
\theoremstyle{remark}
\newtheorem{rem}[thm]{\protect\remarkname}
\theoremstyle{plain}
\ifx\thechapter\undefined
\newtheorem{prop}{\protect\propositionname}
\else
\newtheorem{prop}{\protect\propositionname}[chapter]
\fi
\usepackage{color}
\usepackage{textcomp}
\usepackage{algorithmicx}
\usepackage{algpseudocode}

\usepackage{tikz}
\usetikzlibrary{arrows.meta,positioning}

\theoremstyle{definition}
\newtheorem{definition}{Definition}[section]

\usepackage{babel}

\providecommand{\remarkname}{Remark}
\providecommand{\theoremname}{Theorem}
\providecommand{\propositionname}{Proposition}

\makeatother

\usepackage{babel}
\providecommand{\remarkname}{Remark}
\providecommand{\theoremname}{Theorem}

\begin{document}
\title{Machine Collaboration}
\author{Qingfeng Liu\thanks{Corresponding author. Department of Industrial and Systems Engineering, Hosei University, Koganei City, Tokyo, Japan. Email: qliu@hosei.ac.jp.} \ and Yang Feng\thanks{School of Global Public Health at New York University, NY, NY, USA.
Email: yang.feng@nyu.edu.}}
\maketitle
\begin{abstract}
We propose a new ensemble framework for supervised learning, called
machine collaboration (MaC), using a heterogeneous collection of base
machines for prediction tasks. Unlike bagging/stacking (a parallel
\& independent framework) and boosting (a sequential \& top-down framework),
MaC is a type of\textit{ circular }\&\textit{ interactive} learning
framework. The \textit{circular }\&\textit{ interactive} feature helps
the base machines to transfer information circularly and update their
structures and parameters accordingly. The theoretical result on the
risk bound of the estimator from MaC reveals that the \textit{circular
}\&\textit{ interactive} feature can help MaC reduce risk via a parsimonious
ensemble. We conduct extensive experiments on MaC using both simulated
data and $119$ benchmark real datasets. The results demonstrate that
in most cases, MaC performs significantly better than several other
state-of-the-art methods, including classification and regression
trees, neural networks, stacking, and boosting. 
\end{abstract}

\section{Introduction\label{sec:intro}}

In recent decades, various learning methods, including deep neural
networks (DNN), decision trees (DT, \citet{breiman1984classification}),
support vector machines \citet{cortes1995support}, and k-nearest
neighbors (kNN) have been developed for regression in supervised learning.
As argued by \citet{hastie2009elements}, each of these methods may
have advantages over the others in some respects, but not in others.
For example, while DNN is effective at approximating complicated nonlinear
functions, the problems of overfitting and vanishing gradients could
harm their performances , especially when the sample size is small.
As another popular learning method, DT is robust to irrelevant predictor
variables and outliers and insensitive to monotone transformations
of the input data. However, the lack of smoothness of the prediction
surface is one limitation of DT. Another drawback of DT is that a
slight change in data can result in quite different splits of the
DT, leading to a potentially large variance of prediction. 
Ridge regression is also a popular learning method, being robust against
the multicollinearity of the predictors. However, it can only capture
the linear relationship between the response and predictors. There
is no single learning method that will dominate all others in all
scenarios. In this article, we propose a new ensemble learning framework
to combine the strengths of these different learning methods through
collaboration.

Ensemble learning has emerged and been extensively studied by many
in the past few decades (e.g., \citet{dasarathy1979composite}, \citet{Schapire_1990},
\citet{ho1995random}, and \citet{breiman1996bagging}), with its
popularity recently skyrocketing (e.g., \citet{NIPS2017_49ad23d1},
\citet{yu2018semi}, \citet{NEURIPS2019_3f7bcd0b}, and \citet{tian2021rasea}).
\citet{mendes2012ensemble}, \citet{sagi2018ensemble}, and \citet{dong2020survey}
are some recent comprehensive surveys. The general idea of ensemble
learning is to combine the predictions obtained from different learning
methods (hereafter, base machines), or predictions based on different
subsamples or different feature spaces, in order to improve prediction
performance. Bagging, stacking, and boosting are three prominent examples.
In bagging \citep{breiman1996bagging} and stacking, base machines
are first run in parallel and independently, and then the final prediction
is constructed as a simple/weighted average of the predictions from
these base machines. In boosting \citep{Schapire1998boosting}, the
base machines work jointly in a top-down manner. In all three algorithms,
the output from each base machine is fixed after being calculated.
Like human collaboration, an idea that may yield potential improvement
is to let different kinds of base machines communicate with each other
and update their outputs after observing the predictions of the other
base machines. Based on this idea, we propose the \emph{Machine Collaboration}
(MaC) ensemble learning framework with heterogeneous base machines,
where the word heterogeneous stands for that the base machines are
of different types (e.g. DT, DNN, Ridge Regression). 

Compared with bagging, stacking, and boosting, MaC has the following
desirable features. Figure \ref{fig:Bagging,-boosting-and} provides
the schematic for bagging, stacking, boosting, and MaC. As illustrated,
bagging and stacking are parallel \& independent, boosting is sequential
\& top-down, while MaC is \textit{ circular }\&\textit{ recursive}.
In the framework of MaC, base machines work in a circular manner.
Further, the circulation goes multiple rounds. Valuable information
is passed recursively through base machines around a ``round table",
but not top-down. In this process, the base machines update their
structures and/or parameters once in each round, according to the
information received from the other machines. We demonstrate that
MaC can deliver competitive performance when compared with the base
machines or other ensemble methods.

Three existing ensemble learning methods are closely related to MaC.
One is the so-called super learner (hereafter, SL) proposed in \citet{vanderLaanPolleyHubbard2007},
which can be viewed as a stacking method. SL receives predictions
from different kinds of base machines, then outputs a weighted average
of these predictions as the final prediction, where the weights are
obtained using cross-validation with some specified loss function.
The second one is the LS-Boost proposed in \citet{friedman2001greedy}.
LS-Boost works as a forward regression, using a particular base machine
(e.g. tree) each time to fit the current residuals, and ensemble all
machines in the end. LS-Boost is then a sequential \& top-down method.
While both ensemble methods work well in certain situations, we show
that our newly proposed MaC performs better in our extensive simulation
studies and real data analysis, partly due to its distinctive circular
\& recursive learning structure. The third one is the so-called additive
groves which were proposed to train an additive ensemble of regression
trees (See \citet{sorokina2007additive} and \citet{sorokina2008detecting}).
The main difference is that the additive groves are meant to build
an ensemble of homogeneous models (regression trees), more in the
spirit of LS-boost or bagging, while MaC focuses on heterogeneous
ensembles in the spirit of stacking.

The main contributions of this work are fourfold. First, we propose
a new type of ensemble learning framework, MaC, which is \textit{circular
}\&\textit{ recursive}. The \textit{circular }\&\textit{ recursive}
aspect could be a potential direction for exploring new methods of
ensemble learning. Second, we present some desirable finite statistical
properties of MaC. Third, we demonstrate via extensive simulations
that MaC performs better than all individual base machines and the
ensemble methods SL and LS-Boost. Lastly, in the real data analysis,
we compare MaC with the competing methods on $119$ benchmark datasets
in the Penn Machine Learning Benchmarks (PMLB) \citep{Olson2017PMLB},
which demonstrate notable advantages of MaC over competing methods
for most datasets.

\section{Method\label{sec:Method}}

We now introduce the details of the new ensemble learning framework,
machine collaboration (MaC). In MaC, we consider a collection of base
machines and allow them to collaborate to improve the prediction performance.
The base machines could contain both hyperparameters and non-hyperparameters.
Here, a hyperparameter is a parameter whose value controls the structure
of the base machine and could be set using domain knowledge or selected
by cross-validation or data-splitting. The remaining parameters, whose
values are estimated by fitting the base machine with fixed hyperparameters
on the training data, are non-hyperparameters. For example, in DNN,
the learning rate, number of nodes, number of layers, and the activation
function are hyperparameters, while for DT, the maximum depth is a
hyperparameter. For DNN, the weights of each layer are non-hyperparameters,
and for DT, the split variable and value for each split are non-hyperparameters.
Hereafter, for the sake of simplicity, we use ``parameter'' instead
of ``non-hyperparameter'' unless confusion arises. First, we provide
a simple sketch of MaC with two base machines in the next subsection.
Then the general MaC framework is provided in subsection \ref{subsec:Algorithm-of-general}.

\subsection{Sketch of MaC with two base machines}

\begin{figure}[h]
\begin{centering}
\includegraphics[width=0.7\textwidth]{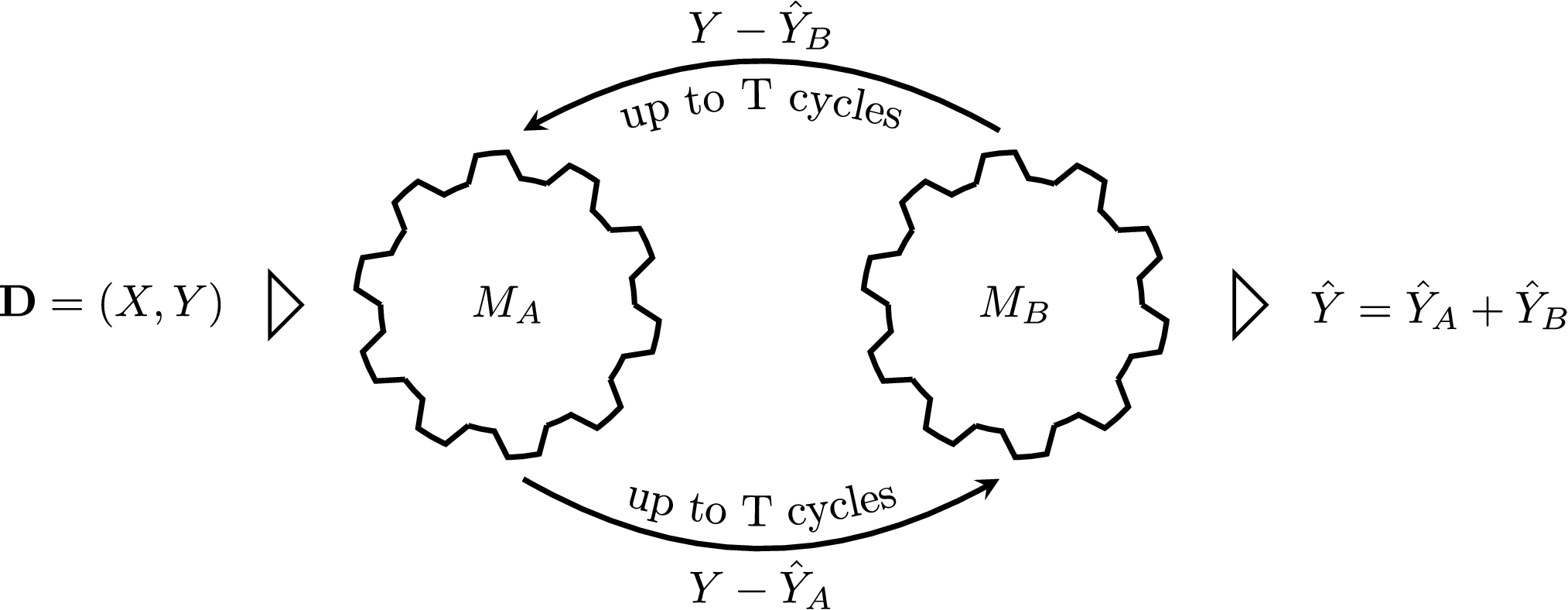} 
\par\end{centering}
\caption{\label{fig:M2}Machine Collaboration}
\end{figure}

Suppose we have a regression task and two base machines, $M_{A}$
and $M_{B}$ (e.g., a DT and a DNN). Denote the dataset as $\mathbf{D}\ensuremath{=}\left\{ \text{feature }\ensuremath{X},\text{ target }Y\right\} $.
For two real-valued $n_{z}$-dimensional vectors $Z=\left(z_{1},\cdots,z_{n_{z}}\right)$
and $\hat{Z}=\left(\hat{z}_{1},\cdots,\hat{z}_{n_{z}}\right)$, denote
the loss function as $L\left(z,\hat{z}\right)$, the empirical risk
function $R(Z,\hat{Z})\equiv\frac{1}{n_{z}}\sum_{j=1}^{n_{z}}L\left(z_{j},\hat{z}_{j}\right)$.
Let $\hat{Y}_{A}$ and $\hat{Y}_{B}$ denote the currently fitted
values from machines $M_{A}$ and $M_{B}$, respectively. Note that
the key idea of MaC is to update $\hat{Y}_{A}$ and $\hat{Y}_{B}$
alternatively throughout the collaboration process. Figure \ref{fig:M2}
is a schematic of MaC with two base machines with the detailed steps
as follows.
\begin{description}
\item [{Step~1}] Randomly split data $\mathbf{D}$ to training data $\mathbf{D}^{t}=\left\{ X^{t},Y^{t}\right\} $
and validation data $\mathbf{D}^{v}=\left\{ X^{v},Y^{v}\right\} $.
Initialize $\hat{Y}=\hat{Y}_{A}=\hat{Y}_{B}=\mathbf{0}$. 
\item [{Step~2}] Update the working response for machine $M_{A}$ as $Y_{A}\equiv Y-\hat{Y}_{B}$.
Construct $\ensuremath{\mathbf{\tilde{D}}=}\left\{ \ensuremath{X},Y_{A}\right\} $
and split it into $\mathbf{\tilde{D}^{t}}$ and $\mathbf{\tilde{D}^{v}}$
accordingly. Tune the hyperparameters and estimate the parameters
of $M_{A}$ with the data $\mathbf{\tilde{D}}$, update the predicted
value $\hat{Y}_{A}$ of $Y_{A}$ using machine $M_{A}$. Then, update
$\hat{Y}=(\hat{Y}^{t},\hat{Y}^{v})\equiv\hat{Y}_{A}+\hat{Y}_{B}$,
where $\hat{Y}^{t}$ and $\hat{Y}^{v}$ denote the predictions of
the training sample and the validation sample, respectively. Calculate
the empirical risk of validation data $R^{v}=R(Y^{v},\hat{Y}^{v})$. 
\item [{Step~3}] Update the working response for machine $M_{B}$ as $Y_{B}\equiv Y-\hat{Y}_{A}$.
Construct $\ensuremath{\mathbf{\tilde{D}}=}\left\{ \ensuremath{X},Y_{B}\right\} $
split it into $\mathbf{\tilde{D}^{t}}$ and $\mathbf{\tilde{D}^{v}}$.
Tune the hyperparameters and estimate the parameters of $M_{B}$ with
the data $\mathbf{\tilde{D}}$, update the predicted value $\hat{Y}_{B}$
of $Y_{B}$ using machine $M_{B}$. Then, calculate $\hat{Y}=(\hat{Y}^{t},\hat{Y}^{v})\equiv\hat{Y}_{A}+\hat{Y}_{B}$
and the empirical risk of validation data $R^{v}=R(Y^{v},\hat{Y}^{v})$. 
\item [{Step~4}] Iterate Steps 2 and 3 with $M_{A}$ and $M_{B}$ up to
$T>0$ times. During each iteration, check the loss of validation
data and stop the iteration if $R^{v}$ does not decrease any more. 
\item [{Step~5}] The final prediction of $Y$ is the $\hat{Y}$ from the
iteration with the smallest $R^{v}$. 
\end{description}
Note that we adopt a hold-out method to evaluate different machines.
In theory, a cross-validation method (hear after, CV), e.g. $k$-fold
CV scheme could be adopted, for our experiments, however, Optuna,
the hyperparameter optimization software framework taken by us prevents
us to adopt a CV method. Taking the $k$-fold CV as an example, Optuna
will not give us the evaluation values for the same hyperparameters
for all the $k$ folds so that we can't take the average of the evaluation
values across the $k$-fold. Developing a new optimization framework
for $k$-fold is an interesting future work.

\subsection{A general algorithm for MaC\label{subsec:Algorithm-of-general}}

We can easily extend the idea in the MaC for two machines to a situation
with more machines. To describe the general algorithm (Algorithm 1)
of MaC with more than $2$ base machines, we need the following setup
and notations. Suppose we have an independent and identically distributed
sample of size $n$ 
\[
\mathbb{\mathbf{D}}=\left(D_{1},\cdots,D_{n}\right)=\left(\left(X_{1},Y_{1}\right),\cdots,\left(X_{n},Y_{n}\right)\right)
\]
generated from the true distribution $P_{0}$, which is an element
of a statistical model $\mathbb{P}$. The support of $P_{0}$ is $\mathcal{D}\equiv\mathbb{\mathcal{X}}\times\mathcal{Y}=\left\{ d\in\mathcal{\mathbb{R}}^{l}\times\mathbb{R}|P_{0}\left(d\right)\neq0\right\} $,
where $l$ is a positive integer denoting the dimension of $\mathbb{\mathcal{X}}$.
MaC is constructed based on $K_{n}$ different base machines $\{m_{k,\lambda_{k},\theta_{k}}\}_{k=1,\cdots,K_{n}}$,
where $\lambda_{k}$ and $\theta_{k}$ are the vectors of hyperparameters
and parameters of the $k$th machine, respectively. Here, we use $K_{n}$
to allow the number of base machines $K$ to grow alongside sample
size $n$, as demonstrated in Section \ref{sec_Theory}. We use the
same loss function and risk function for tuning the parameters and
hyperparameters of the base machines and the MaC throughout Algorithm
1. Moreover, assume the tuning and estimation algorithm for each base
machine is given. Let $\hat{Y}_{k}^{t}$ and $\hat{Y}_{k}^{v}$ denote
the predictions of the training sample and validation sample based
on the $k$th machine, respectively. For machine $k=1,\cdots,K_{n}$,
define $\hat{Y}_{k}\equiv(\hat{Y}_{k}^{t},\hat{Y}_{k}^{v})$, $\hat{Y}_{-k}\equiv\sum_{j\neq k}^{K_{n}}\hat{Y}_{j}$
and $\hat{Y}^{v}\equiv\sum_{j=1}^{K_{n}}\hat{Y}_{j}^{v}$. Define
the index pair of the outer and inner loops when the loop stops as
$\left\{ i^{*},k^{*}\right\} $. 
\begin{algorithm}
	\SetKwProg{Fn}{Def}{\string:}{}
	\SetAlgoLined
	\SetKwInput{KwRequire}{Require}
	\KwRequire{$K>1$ different base machines $\{m_{k,\lambda_{k},\theta_{k}}\}_{k=1, \cdots, K}$. Maximum Tolerance integers $\tau > 0$ and $T > 0$.}
	\SetKwInput{KwDefination}{Define}
	\KwIn{$\mathbf{D}=\left\{\text{feature data }X,\text{ target data }Y\right\}$, a new feature data $X^{*}\sim P_0$. }
	\KwOut{The trained MaC, $\hat{\mathfrak{m}}:\mathcal{X}\rightarrow \mathcal{Y}$; the prediction based on $X^*$, $\hat{Y}^{*}$.} 
{
	Randomly split $\mathbf{D}$ into training data $\mathbf{D}^t\equiv(X^t,Y^t)$ and validation data $\mathbf{D}^v\equiv(X^v,Y^v)$\ so that the proportion of the validation data is $p$;
	
	Initialize $t=0$, $R_0=\infty$, $\hat{Y}_k=(\hat{Y}_k^t,\hat{Y}_k^v)=(\mathbf{0},\mathbf{0})$ for $k =1,\cdots,K$;
	
	\For{$i=1$ \KwTo $T$}{
		
		\For{$k=1$ \KwTo $K$}{
			
			Construct $\tilde{\mathbf{D}}^t$ and $\tilde{\mathbf{D}}^v$ by replacing $Y$ in $\mathbf{D}^t$ and $\mathbf{D}^v$ with $Y_k \leftarrow ( Y-\hat{Y}_{-k} )$, where $\hat{Y}_{-k}=\sum_{j\neq k}^{K}\hat{Y}_{j}$;

			

			For $m_{k,\lambda_{k},\theta_{k}}$, tune the hyperparameters and estimate the non-hyperparameters, obtain $\hat{\lambda}_{k}^{(i)}$ and $\hat{\theta}_{k}^{(i)}$, using $\tilde{\mathbf{D}}^t$ and $\tilde{\mathbf{D}}^v$ as traning data and validation data, respectively.

			$\hat{Y}_{k}=\left(\hat{Y}_{k}^{t},\hat{Y}_{k}^{v}\right)\leftarrow\left(m_{k,\hat{\lambda}_{k}^{(i)},\hat{\theta}_{k}^{(i)}}\left(X^{t}\right),m_{k,\hat{\lambda}_{k}^{(i)},\hat{\theta}_{k}^{(i)}}\left(X^{v}\right)\right)$;
			
			$\hat{Y}^v=\sum_{j=1}^{K}\hat{Y}_{j}^{v}$;
			$R_{k}^{v} \leftarrow R(Y^v,\hat{Y}^v)$;
			
		}
		
		$\tilde{k}=\arg\!\min_{k}R_k^v$;
		
		\SetKwFor{If}{if}{}{}
		\If{$R^v_{\tilde{k}} < R_0$}{$R_0 \leftarrow R^v_{\tilde{k}}$; $k^{'}\leftarrow\tilde{k}$; $i^{'}\leftarrow i$; $t = 0$;}
		
		\SetKwFor{Else}{else}{}{end else}
		\Else{}{$t\,+\!=1$;}
		\SetKwFor{If}{if}{}{end}
		\If{$t \geq \tau$ or $i \geq T$}{
			$i^{*}=i^{'}$; $k^{*}=k^{'}$;
			
			\textbf{break};}
	}
	\KwRet{
		\vspace*{-10pt}
		\[
		\begin{array}{cc}
			\begin{array}{cl}
				\hat{\mathfrak{m}}= & \left\{  m_{1,\hat{\lambda}_{1}^{(i^{*})},\hat{\theta}_{1}^{(i^{*})}}+\cdots+m_{k^{*},\hat{\lambda}_{k^{*}}^{(i^{*})},\hat{\theta}_{k^{*}}^{(i^{*})}}\right.\\
				& +\left.m_{k^{*}+1,\hat{\lambda}_{k^{*}+1}^{(i^{*}-1)},\hat{\theta}_{k^{*}+1}^{(i^{*}-1)}}+\cdots+m_{K,\hat{\lambda}_{K}^{(i^{*}-1)},\hat{\theta}_{K}^{(i^{*}-1)}}
				\right\},
			\end{array} & \hat{Y}^*=\hat{\mathfrak{m}}\left(X^*\right);\end{array}
		\]
	}
}
\caption{Algorithm for Machine Collaboration (MaC)}
\end{algorithm}

When the sample size is not large, if we do not use the full data
but only $\mathbf{D}^{t}$ to estimate the non-hyperparameter vector,
there could be a large information loss. To avoid this, in our experiments
in Section \ref{sec_Experiments}, we follow common practice in machine
learning (See Listing 4.1 in p.98 of \citet{chollet2017deep}), namely
to use the full data to estimate the non-hyperparameter vector in
step 6 of Algorithm 1. Note that, there is a trade-off between computation
cost and the accuracy of prediction. To get a higher accuracy of prediction,
you would need bigger $T$ or $K_{n}$, leading to higher computation
cost.

\section{Theory for finite sample}

\label{sec_Theory}

Under the setting of subsection \ref{subsec:Algorithm-of-general},
denote the cumulative distribution function of $X_{1}$ as $F_{0}$.
Let $B_{n}\in\left\{ 0,1\right\} ^{n}$ be a random binary $n$-vector
whose observed value defines a split of the data $\mathbf{D}$ into
a training sample $\mathbf{D}^{t}$ and a validation sample $\mathbf{D}^{v}$,
with $1$ for validation and $0$ for training. Let $p$ denote the
proportion of observations in the validation sample, and $P_{n}$,
$P$$_{n,B_{n}}^{t}$ and $P_{n,B_{n}}^{v}$ denote the empirical
distributions of $\mathbf{D}$, $\mathbf{D}^{t}$, and $\mathbf{D}^{v}$,
respectively.

Suppose we have a set of $K_{n}$ base machines $\left\{ M_{1},\cdots,M_{K_{n}}\right\} $.
Assume the space of the hyperparameter $\lambda_{j}$ and the parameter
vector $\theta_{j}$ of the $j$th base machine are $\Lambda_{j}\subseteq\mathbb{R}^{d_{\lambda,j}}$
and $\Theta_{j}\subseteq\mathbb{R}^{d_{\theta,j}}$, respectively.
Then, each base machine $M_{j}:\mathbb{P}\rightarrow\mathcal{S}_{j}\left(\mathcal{X}|\Lambda_{j}\times\Theta_{j}\right)$
is a mapping from $\mathbb{P}$ into $\mathcal{S}_{j}\left(\mathcal{X}\right)\equiv\mathcal{S}_{j}\left(\mathcal{X}|\Lambda_{j}\times\Theta_{j}\right)$.
$\mathcal{S}_{j}\left(\mathcal{X}|\Lambda_{j}\times\Theta_{j}\right)$
is a space of real-valued parametric functions from $\mathcal{X}$
to $\mathbb{R}$, taking the vectors of the hyperparameters and parameters
in the space $\Lambda_{j}\times\Theta_{j}$. For $j=1,\cdots,K_{n}$,
denote the realization of $M_{j}$ as $m_{j,\lambda_{j},\theta_{j}}:\mathcal{X}\rightarrow\mathbb{R}$,
which is a function with hyperparameter vector $\lambda_{j}\in\Lambda_{j}$
and parameter vector $\theta_{j}\in\Theta_{j}$.

We consider candidate MaCs constructed by the sum of $K_{n}$ base
machines with certain hyperparameters and parameters. In particular,
we have $\mathcal{M}\equiv M_{1}(\cdot)+\cdots+M_{K_{n}}(\cdot):\mathbb{P}\rightarrow\mathcal{S}(\mathcal{X})\equiv\mathcal{S}\left(\mathcal{X}|\Lambda(n)\times\Theta(n)\right)$,
where $\Lambda(n)=\prod_{j=1}^{K_{n}}\Lambda_{j}$ and $\Theta(n)=\prod_{k=1}^{K_{n}}\Theta_{k}$
denotes the space of the hyperparameter vector $\lambda\equiv\left\{ \lambda_{1},\lambda_{2}\cdots,\lambda_{K_{n}}\right\} $
and parameter vector $\theta\equiv\left\{ \theta_{1},\theta_{2},\cdots,\theta_{K_{n}}\right\} $
of MaC, respectively.

Define the space of all candidate MaCs as $\mathbb{M}\equiv\left\{ \mathcal{M}\left(P\right):P\in\mathbb{P}\right\} \subseteq\mathcal{S}(\mathcal{X})$,
then each realization of $\mathcal{M}$, $\mathfrak{m}\equiv\mathfrak{m}_{\lambda,\theta}:\mathcal{X}\rightarrow\mathbb{R}$
is an element of $\mathbb{M}$ with hyperparameter vector $\lambda\in\Lambda(n)$
and parameter vector $\theta\in\Theta(n)$. Denote the loss function
of $m$ by $L\left(D,m\right)$ and the risk of $m$ by $E_{P_{0}}L\left(D,m\right)$
for $D\in\mathcal{D}$, and $m$ denoting a base machine or a MaC.
Endow the space $\mathcal{S}(\mathcal{X})$ a dissimilarity function
$\delta:\mathcal{S}(\mathcal{X})\times\mathcal{S}(\mathcal{X})\rightarrow\mathbb{R}$,
with the dissimilarity between $\mathfrak{m}_{1}$ and $\mathfrak{m}_{2}$
in $\mathbb{M}$ defined as 
\[
\delta\left(\mathfrak{m}_{1},\mathfrak{m}_{2}\right)\equiv\int\left|L(d,\mathfrak{m}_{1})-L\left(d,\mathfrak{m}_{2}\right)\right|\mathrm{d}P_{0}(d).
\]
Define the pseudo-true MaC as 
\begin{align*}
\mathfrak{m}_{0}\equiv\mathcal{M}\left(P_{0}\right)= & \arg\!\min{}_{\mathfrak{m}\in\mathbb{M}}E_{P_{0}}L(D,\mathfrak{m})\\
= & \arg\!\min{}_{\mathfrak{m}\in\mathbb{M}}\int L(d,\mathfrak{m})\mathrm{d}P_{0}(d),
\end{align*}

and the risk difference of $\mathfrak{m}$ as $\delta\left(\mathfrak{m},\mathfrak{m}_{0}\right)$.

Denote the space of the hyperparameter vector in which the MaC algorithm
searches by $\tilde{\Lambda}(n)\subseteq\Lambda(n)$ with cardinality
$\left|\tilde{\Lambda}(n)\right|=J_{n}$. Indexing the elements of
$\tilde{\Lambda}(n)$ as $\left\{ \lambda_{\left(1\right)},\cdots,\lambda_{\left(J_{n}\right)}\right\} $,
we construct $J_{n}$ subspace $\mathbb{M}_{k}\equiv\left\{ \mathfrak{m}_{\lambda_{\left(k\right)},\theta}:\theta\in\Theta(n)\right\} \subseteq\mathbb{M}$
according to the selection of hyperparameter vector for $k\in\kappa_{n}\equiv\left\{ 1,\cdots,J_{n}\right\} $.
For a particular subspace $\mathbb{M}_{k}$ with fixed hyperparameter
vector $\lambda_{\left(k\right)}$, the MaC algorithm searches for
the optimal parameter vector on the parameter space $\tilde{\Theta}_{k}(n)\subseteq\Theta(n)$.
After collecting MaCs whose hyperparameter vector is $\lambda_{\left(k\right)}$
and parameter vector $\theta\in\tilde{\Theta}_{k}(n)$, we create
the space $\mathbb{M}_{k,\tilde{\Theta}_{k}}\equiv\left\{ \mathfrak{m}_{k,\theta}\equiv\mathfrak{m}_{\lambda_{\left(k\right)},\theta}:\theta\in\tilde{\Theta}(n)\right\} \subseteq\mathbb{M}_{k}$.
Define the risk approximation error of $\mathbb{M}_{k,\tilde{\Theta}_{k}}$
as $B_{0}\left(k\right)\equiv\min_{\mathfrak{m}\in\mathbb{M}_{k,\tilde{\Theta}_{k}}}\delta\left(\mathfrak{m},\mathfrak{m}_{0}\right)$.
Note that, in van der Laan (2006), $\mathbb{M}_{k,\tilde{\Theta}_{k}}$
is called epsilon-net, which is a form of sieving net for the parameters.

For any empirical distribution $P_{n}$, define the estimated model
with fixed $k$ based on our algorithm as 
\[
\hat{\mathcal{M}}_{k}\left(P_{n}\right)\equiv\underset{\mathfrak{m}\in\mathbb{M}_{k,\tilde{\Theta}_{k}(n)}}{\arg\!\min}\int L(d,\mathfrak{m})\mathrm{d}P_{n}(d).
\]
Then, the estimated MaC is $\hat{\mathcal{M}}\left(P_{n}\right)\equiv\hat{\mathcal{M}}_{k\left(P_{n}\right)}\left(P_{n}\right),$
where $k\left(P_{n}\right)\equiv\arg\!\min_{k\in\kappa_{n}}\int L(d,\hat{\mathcal{M}}_{k}\left(P_{n}^{t}\right))\mathrm{d}P_{n}^{v}(d)$.

To introduce the following theorem, we need the following definition.
\theoremstyle{definition} \begin{definition}[Searching Number
and Searching Resolution]Let $\tilde{\Theta}_{k}(n)=\left\{ \theta_{\left(k,1\right)},\cdots,\theta_{\left(k,N_{k}\right)}\right\} $
with $N_{k}<\infty$. For a real number $\varepsilon>0$, define a
sphere $B\left(\mathfrak{m}_{j},\varepsilon\right)\equiv\left\{ \mathfrak{m}\in\mathbb{M}_{k}:\left|\mathfrak{m}-\mathfrak{m}_{j}\right|\leq\varepsilon\right\} $.
We refer to $N_{k}$ as the searching number and $\varepsilon_{k}\equiv\inf_{\varepsilon}\left\{ \varepsilon:\mathbb{M}_{k}\subseteq\cup_{j=1}^{N_{k}}B\left(\mathfrak{m}_{\left(k,j\right)},\varepsilon\right)\right\} $
as the searching resolution of $\mathbb{M}_{k,\tilde{\Theta}_{k}}$
for the algorithm, where $\mathfrak{m}_{k,j}$ denotes the MaC in
$\mathbb{M}_{k}$ with the non-hyperparameter vector $\theta_{\left(k,j\right)}$.

\end{definition}

Note that in the definition above, the space $\tilde{\Theta}_{k}(n)$
is first fixed. If we first fix $\varepsilon_{k}$, and then select
a space $\tilde{\Theta}_{k}(n)$ with minimum $N_{k}$ so that $\cup_{j=1}^{N_{k}}B\left(\mathfrak{m}_{\left(k,j\right)},\varepsilon\right)$
can cover $\mathbb{M}_{k}$, then $N_{k}$ is just what we would call
a covering number.

Let us refer to the value of the parameter vector that results in
the minimum of the empirical loss of estimation as the optimal parameter
vector. {In the theoretical analysis, following \citet{vanderLaanPolleyHubbard2007},
we only consider a MaC algorithm that searches the optimal parameter
vector on a discrete set.\footnote{{[}}1{]}{{As $N_{k}$ goes to
infinity with $n$, Theorem \ref{Thm:1} can be regarded as an approximate
result for the MaC algorithm that searches the optimal parameter vector
on the whole domain.}} For $\tilde{\Theta}_{k}(n)=\left\{ \theta_{\left(k,1\right)},\cdots,\theta_{\left(k,N_{k}\right)}\right\} $
with $N_{k}<\infty$, we have the following finite sample result.} 
\begin{thm}
\label{Thm:1}Assume a constant $C_{0}<\infty$ exists, $\left|Y\right|\leq C_{0}$
$a.s.$ and $\sup_{\mathfrak{m}\in\mathbb{M}}\sup_{X\in\mathcal{X}}\left|\mathfrak{m}\left(X\right)\right|\leq C_{0}$.
Define $C_{1}\equiv4C_{0}^{2}$ and $C_{2}\equiv16C_{0}^{2}$. Let
$L\left(D,\mathcal{\mathfrak{m}}\right)\equiv\left(Y-\mathfrak{m}\left(X\right)\right)^{2}$,
$\mathfrak{m}_{0}\left(X\right)\equiv E_{P_{0}}\left[Y|X\right]$,
$C(a)\equiv4(1+a/2)^{2}\left(\frac{2C_{1}}{3}+\frac{2C_{2}}{a}\right)$.
Then, for any $a>0$ the following inequality holds 
\begin{align}
E_{P_{0}} & \delta\left(\hat{\mathcal{M}}\left(P_{n,B_{n}}^{t}\right)(x),\mathfrak{m}_{0}(x)\right)\nonumber \\
\leq & (1+a)\times\min_{k\in\kappa_{n}}\left\{ (1+a)\tilde{B}_{0}\left(k\right)\right.\label{eq:reg}\\
 & \left.+C(a)\frac{1+\log\left(N_{k}\right)}{n(1-p)}\right\} +C(a)\frac{1+\log\left(J_{n}\right)}{np},\nonumber 
\end{align}
where $\tilde{B}_{0}\left(k\right)=\min_{\mathfrak{m}\in\mathbb{M}_{k,\tilde{\Theta}_{k}}}\int\left(\mathfrak{m}(x)-\mathcal{\mathfrak{m}}_{0}(x)\right)^{2}dF_{0}(x)$.
\end{thm}

The proof of Theorem \ref{Thm:1} is provided in the supplement. Theorem
\ref{Thm:1} shows that the expectation of the risk difference of
$\hat{\mathcal{M}}\left(P_{n,B_{n}}^{t}\right)(x)$ and $\mathfrak{m}_{0}(x)$
depends on the searching number $N_{k}$, and the cardinality $J_{n}$
of the searching space of the hyperparameters $\tilde{\Lambda}(n)$.
Suppose we search for the optimal MaC on $\mathbb{M}_{k}$. If the
searching number $N_{k}$ of $\mathbb{M}_{k}$ is small, the searching
resolution $\varepsilon_{k}\left(N_{k}\right)$ may be large, and
then $\tilde{B}_{0}\left(k\right)$ could be large too. As a result,
a small $N_{k}$ could result in the risk bound increasing. Moreover,
if $J_{n}$ is small, the value of $\min_{k\in\kappa_{n}}\tilde{B}_{0}\left(k\right)$
could be large. To reduce the prediction risk, we need to not only
adjust $N_{k}$ to strike a balance between the term associated with
$\log\left(N_{k}\right)$ and that associated with risk in the approximation
error $\tilde{B}_{0}\left(k\right)$ but also adjust the balance between
$\min_{k\in\kappa_{n}}\tilde{B}_{0}\left(k\right)$ and the term with
$\log\left(J_{n}\right)$. 
Note that increasing the number of base machines, $K_{n}$, may reduce
the first term of eq.(\ref{eq:reg}), but cause the term with $\log\left(J_{n}\right)$
to increase, hence a proper selection of the base machines and a moderate
$K_{n}$ would also be helpful to reduce the risk.

Theorem \ref{Thm:1} also illustrates a merit of the \textit{circular
}\&\textit{ recursive} feature of MaC, that is, MaC only needs to
add up the final $K_{n}$ machines during the loop of the algorithm
without the need to record all single machines in the loop. Note that
$K_{n}$ is ordinarily much smaller than the total number of all single
machines of the algorithm. Using a \textit{circular }\&\textit{ recursive}
type algorithm can reduce $J_{n}$ compared with a \textit{sequential
}\&\textit{ top-down} type algorithm because a \textit{sequential
}\&\textit{ top-down} type algorithm needs to add all single machines
into its final estimate. 
\begin{rem}
\textit{A related method of MaC is LS-Boost \citep{friedman2001greedy},
which essentially works via repeatedly fitting the residual on the
predictors using one base machine at a time. Recall that we have $K_{n}$
base machines. For MaC, in Algorithm 1, the estimated machine in step
$i$, $m_{k,\hat{\lambda}_{k}^{(i)}\hat{\theta}_{k}^{(i)}}$, is an
updated version of $m_{k,\hat{\lambda}_{k}^{(i-1)}\hat{\theta}_{k}^{(i-1)}}$,
for $i>1$. In the MaC, the machines collaborate with each other by
updating themselves after considering the actions of all others. Note
that we only need to store one set of estimates for each machine at
any given time. The final result of MaC is the summation of the $K_{n}$
machines with their most recent estimates. In contrast, for LS-Boost,
if we expect to modify the result of the first round based on $K_{n}$
estimated base machines, we need to use a special version of LS-Boost,
in which we fit the $K_{n}$ base machines $r_{n}>1$ times in a sequential
fashion. In such an LS-Boost, late-coming machines complement earlier
machines. The results of all single machines are kept and added to
the final result. Finally, the estimated result of LS-Boost is the
summation of the results from the $r_{n}\times K_{n}$ estimated base
machines. As a result, the total number of single machines, $r_{n}\times K_{n}$,
for LS-Boost could be much larger than that, $K_{n}$, for MaC. A
large total number of single machines could increase the risk of LS-Boost.
Admittedly, this is not decisive for a comparison between LS-Boost
and MaC because there are additional factors such as the searching
space of every single machine. As a result, LS-Boost and MaC may have
their own advantages and disadvantages in different situations.} 
\end{rem}

\section{Experiments}

\label{sec_Experiments} 

\subsection{Artificial simulation experiments\label{subsec:sim}}

We first generate data for $10$ independent variables, $x_{1},x_{2},\cdots,x_{10}$,
following a multivariate normal distribution $N(0,\Sigma)$, with
a variance--covariance matrix $\Sigma=[\Sigma_{ij}]_{10\times10}$,
where $\Sigma_{ij}=\rho^{|i-j|}$ with $\rho=0.1$. Then, generate
the error term $\varepsilon\sim N\left(0,1\right)$. We consider the
following two data-generating processes (DGPs). 
\begin{align*}
\textrm{DGP 1: }y & =c_{0}+c_{1}\left(x_{1}+x_{2}+0.5x_{3}+0.3x_{4}+0.2x_{5}\right)^{3}\\
 & +c_{2}I\left(x_{4}>0\right)+c_{3}I\left(x_{5}>1\right)\\
 & +c_{4}I\left(x_{1}x_{2}>0\right)+\varepsilon,\\
\textrm{DGP 2: }y & =c_{0}+c_{1}\left(x_{1}+x_{2}+0.5x_{3}+0.3x_{4}+0.2x_{5}\right)^{3}\\
 & +c_{2}I\left(x_{1}>0\right)\times x_{2}-c_{3}I\left(x_{1}<1\right)\times x_{2}\\
 & +c_{4}I\left(x_{1}>0\right)\times3^{x_{2}}+c_{5}I\left(x_{3}x_{4}>0\right)\\
 & \times\sin\left(x_{5}\right)+\varepsilon,
\end{align*}
where $I(\cdot)$ stands for indicator function and the constants
$c_{0}$ through $c_{5}$ are chosen to standardize each term (mean
0 and variance 1). It is clear to see that DGP 2 has stronger nonlinearity
than DGP 1. Note that the variables $x_{6},\cdots,x_{10}$ are not
included in either of the DGPs, but are included as predictors for
all methods. The sample size $n$ and the number of replications for
both is $1000$. We set $p=0.25$. The sample is randomly split into
training data of $600$ observations, validation data of $200$ observations,
and test data of $200$ observations. We choose the mean squared prediction
error $\textrm{MSPE}_{i}=\frac{1}{n}\sum_{j=1}^{n}\left(\hat{y}_{i,j}-y_{j}\right)^{2}$
as the loss for both the artificial and real data experiments, where
$\hat{y}_{i,j}$ is the prediction of $y_{j}$ for the $i$th replication.

We use DT (CART, \citet{breiman1984classification}), DNN, and Ridge
regression as base machines. The DNN used here has $5$ dense layers.
Each of the first $4$ dense layers is associated with a dropout layer.
We treat the number of nodes of the latent layers, the types of activation
functions, the dropout ratio, the learning rate of optimization, the
batch size, and the number of epochs as hyperparameters. The maximum
depth of the DT is fixed as $10$. The DT is pruned according to the
cost complexity parameter, which is a hyperparameter. The tuning parameter
that controls the strength of the penalty of Ridge is also a hyperparameter.
The order of the base machines is also treated as a hyperparameter.
As mentioned in Section \ref{sec:Method}, all the hyperparameters
are tuned with respect to the MSPE in the validation data. Using these
three base machines for MaC, LS-boost, and SL, we compare the prediction
performances on the test data between MaC, SL, and LS-Boost for both
DGPs. For MaC, we set $T=50$ and $\tau=10$.

We conduct LS-boost with a similar setup of MaC. The three base machines
are repeatedly used with a fixed order for at most $T=50$ times.
The order is selected according to the empirical risk of the validation
data. The processes will be stopped if the empirical risk of the validation
data does not decrease for $\tau=10$ times. For LS-boost and SL,
the hyperparameters of each base machine are tuned according to the
empirical risk of the validation data, when we fit each base machine.

We use the Greene HPC Cluster of New York University to carry out
all the experiments. 
All the experiments are carried out by CPU with a 2x Intel Xeon Platinum
8268 24C 205W 2.9GHz Processor. For each simulation replication, the
computing time of the artificial simulation is about $64$ minutes,
while that of the real data experiments with datasets of different
sizes ranges from $42$ minutes to $50$ hours.

\begin{figure}
	\centering
	\begin{minipage}[c]{0.9\textwidth}
		\subfloat[DGP 1\label{fig:DGP-1hist}]{\noindent \begin{centering}
				\includegraphics[bb=0bp 0bp 452bp 338bp,width=0.5\textwidth]{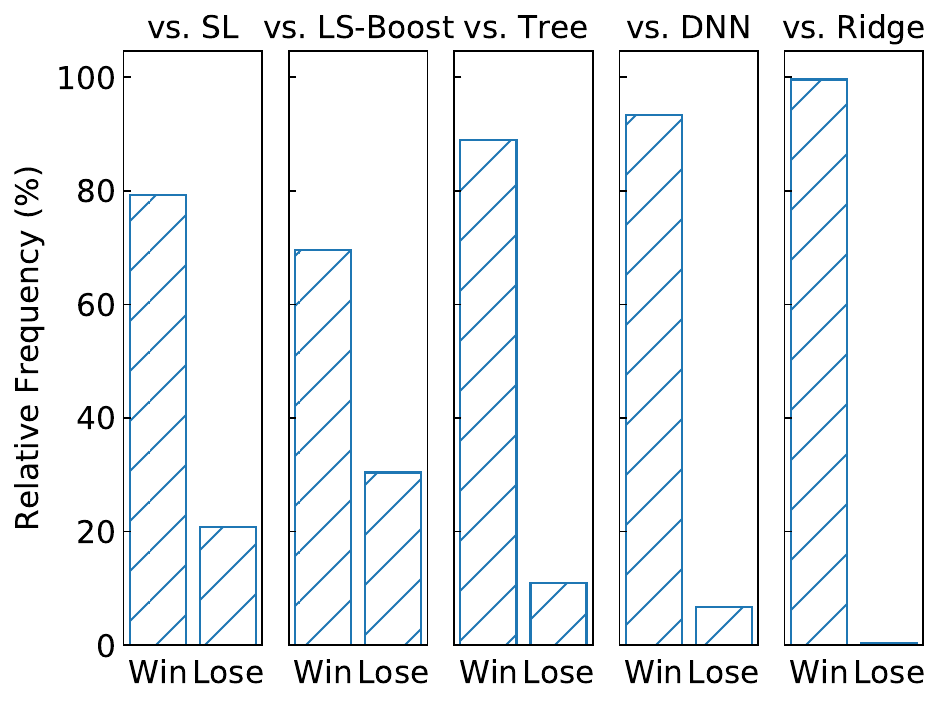}
				\par\end{centering}
		}\hfill{}\subfloat[DGP 1\label{dgp1box}]{\noindent \begin{centering}
				\includegraphics[bb=0bp 0bp 454bp 351.022bp,width=0.5\textwidth]{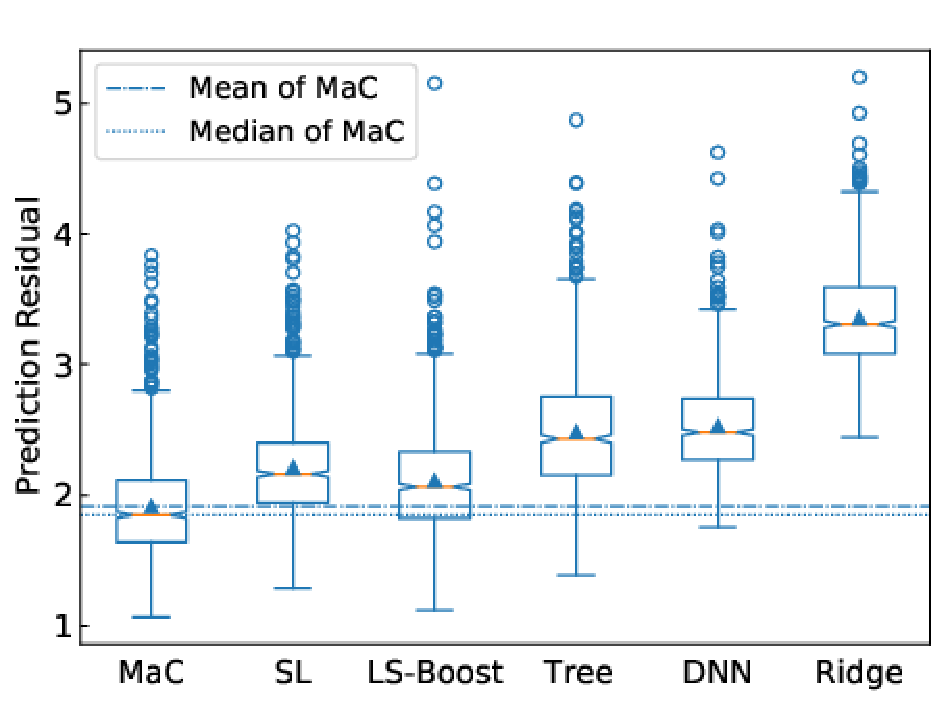}
				\par\end{centering}
		}\vspace{-0.2cm}
		\subfloat[Real Data\label{realdata_hist}]{\begin{centering}
				\includegraphics[bb=0bp 0bp 452bp 338bp,width=0.5\textwidth]{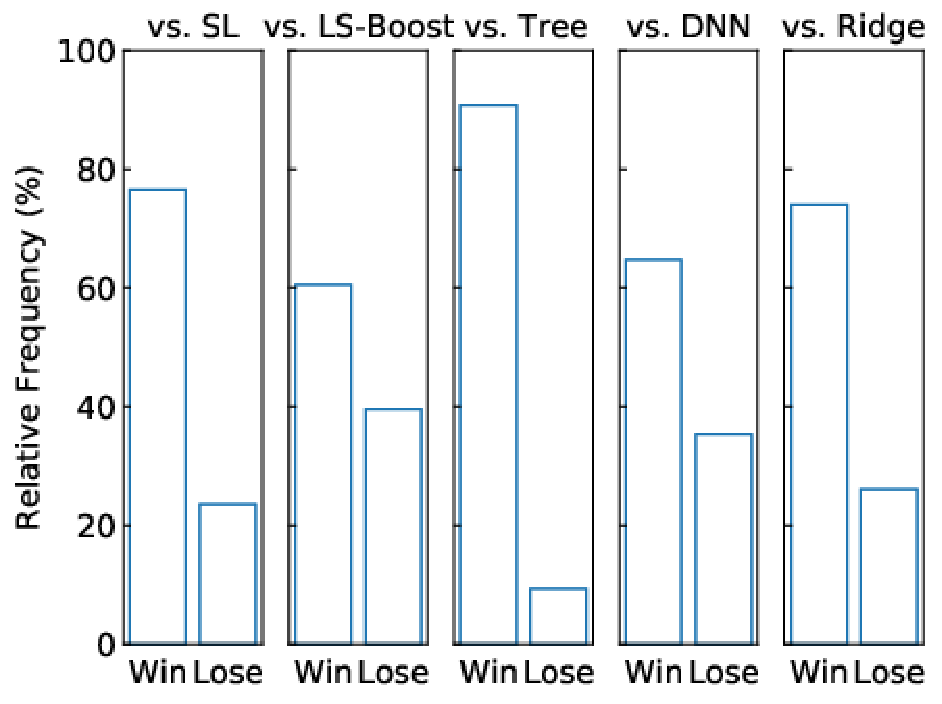}
				\par\end{centering}
		}\hfill{}\subfloat[Real Data\label{realdata_box}]{\noindent \begin{centering}
				\includegraphics[bb=0bp 0bp 454bp 339.698bp,width=0.5\textwidth]{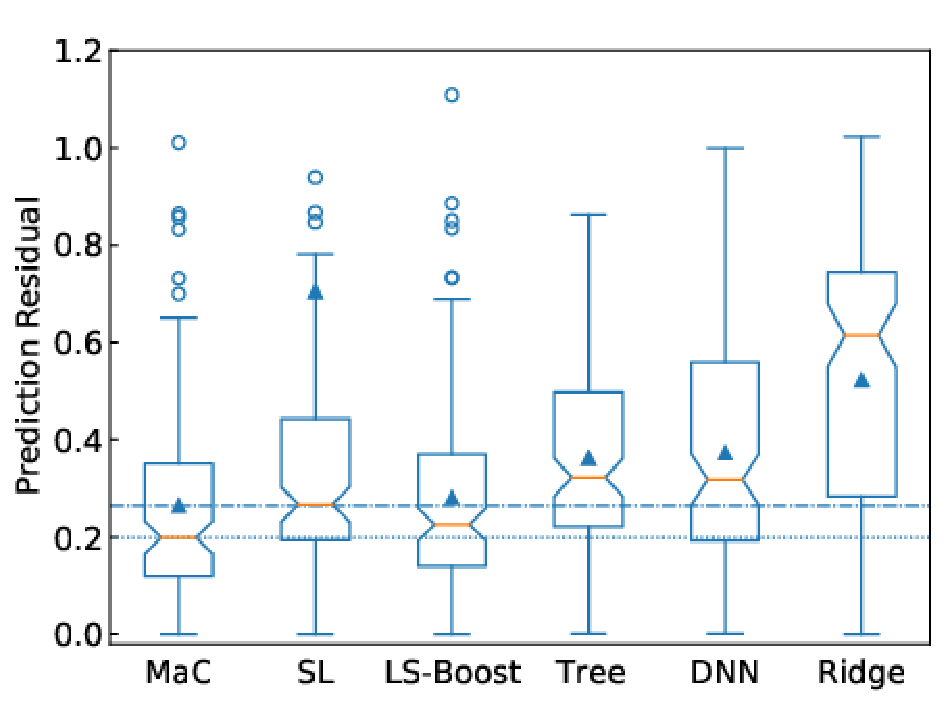}
				\par\end{centering}
		}\caption{MaC vs alternatives}
	\end{minipage}
\end{figure}

The simulation results for DGP 1 are in Figures \ref{fig:DGP-1hist}
and \ref{dgp1box}, with that of DGP 2 in Figures \ref{dgp2hist}
and \ref{dgp2box}. We count the number of replications for which
MaC wins, i.e., it produces a smaller MSPE for the test data than
its competitors, namely a particular ensemble method or a base machine.
As depicted in Figures \ref{fig:DGP-1hist} and \ref{dgp2hist}, for
both DGPs, MaC generally outperforms all the other methods. The boxplots
in Figure \ref{dgp1box} and \ref{dgp2box} show that MaC has a smaller
mean and median of prediction errors on the test data than all the
other methods, whose means and medians are marked by the blue rectangles
and the orange lines, respectively. For the simulation and the real
data experiment (in the next subsection), we calculate the mean and
median of MSPE for each method, the paired t-statistic, and Cohen's
$d$ \citep{cohen1988statistical} of the means of MSPEs for each
pair, which consists of an alternative method and MaC. The results
of the simulations are tabulated in the first two panels of Table
\ref{tbl:sim}. For both DGPs, all the paired t-statistics are large
with values of at least $4.6$, and most of Cohen's $d$s are greater
than $0.2$, implying that the mean and median of MSPE for the MaC
are significantly smaller than those of the other methods.

\subsection{Real data experiment\label{subsec:Realdata}}

\begin{figure}
\centering\includegraphics[width=1\linewidth]{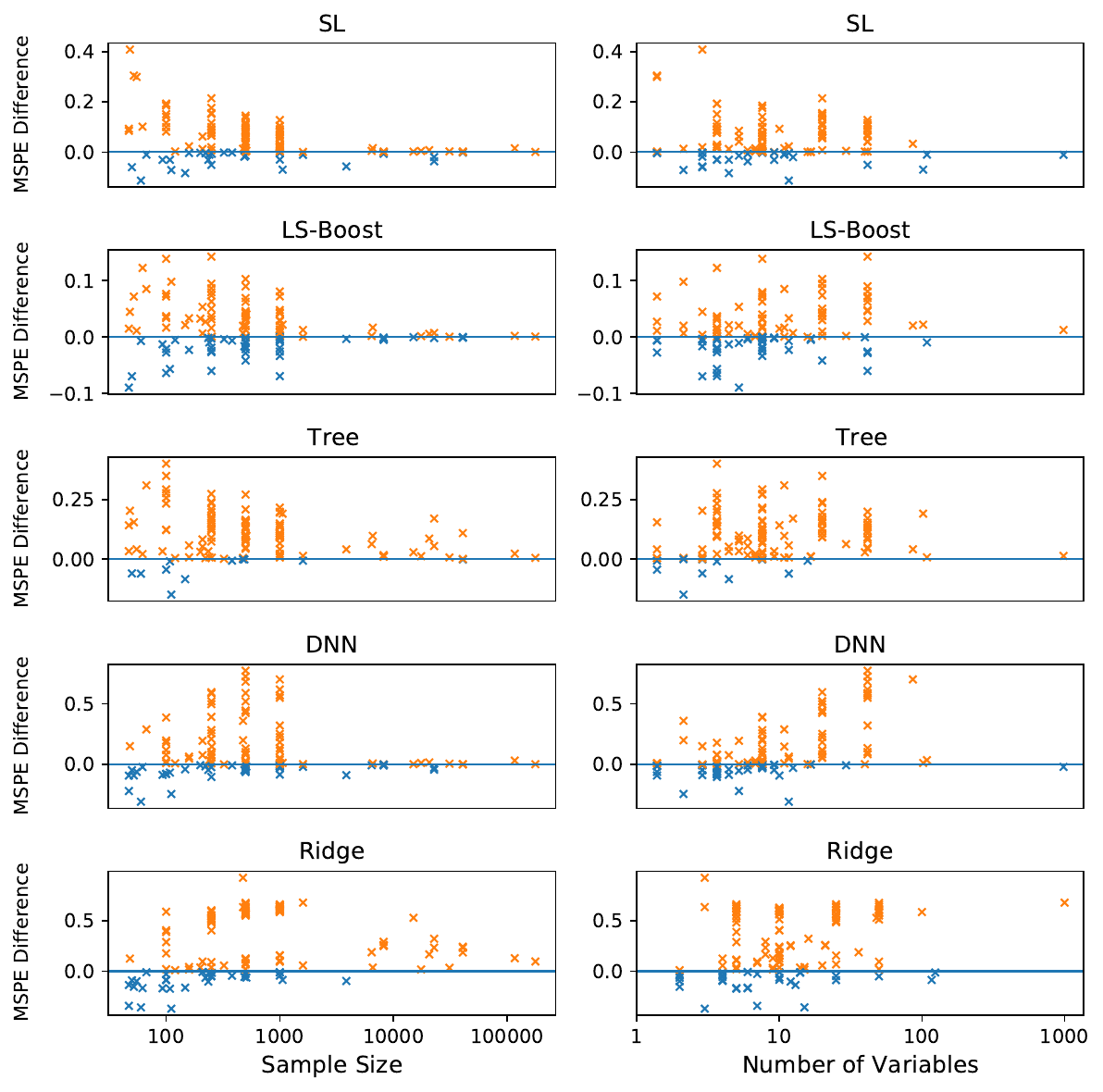}\caption{\label{fig:MSE_difference}MSPE difference ($\mathrm{Alternative}-\mathrm{MaC}$)}
\end{figure}

We use the same base machines and the same setting as the simulations
for our real data experiment. The real datasets are from PMLB, which
is an open-source dataset collection for benchmarking machine learning
methods. All the datasets do not contain personally identifiable information
or offensive content. There are 122 datasets for regression in PMLB
in total, including data about automobile prices, faculty salaries,
pollution, and crime. The sample sizes of the datasets range from
$47$ to $1025010$. We dropped three datasets with sample sizes greater
than or equal to $1$ million because of our limits in computing resources.
As a result, the datasets used have sample sizes ranging from $47$
to $177147$ with a mean of $5476.69$, while the numbers of features
are from $2$ to $1000$ with a mean of $26.05$. To make the comparison
between the different methods simpler, we standardize all variables
in the datasets so they all have a mean of $0$ and a variance of
$1$. 
We apply MaC, SL, LS-Boost, and the three individual base machines
to predict

the target variable in the test data for each dataset. We perform
the experiment $20$ times, in which for each time, each dataset is
randomly split into training ($64\%$), validation ($16\%$), and
test ($20\%$). All the prediction results are in Figures \ref{realdata_hist},
\ref{realdata_box}, and \ref{fig:MSE_difference}. We count the number
of datasets for which MaC outperforms its competitors in terms of
MSPE on the test data. The results are in Figure \ref{realdata_hist},
which shows that MaC wins for more than $60\%$ datasets against all
competitors. Moreover, as illustrated in Figure \ref{realdata_box},
MaC has a smaller mean and median of MSPE than all the competing methods.
To obtain a closer look at the settings where MaC has an advantage,
we calculate the MSPE differences by subtracting the MSPEs of the
other methods from that of MaC. The results of the MSPE differences
are plotted in Figure \ref{fig:MSE_difference} for all the methods.
As shown, for each subfigure, there are more points with positive
MSPE difference than with negative MSPE difference. It means that
MaC has smaller MSPE than corresponding alternative methods. In Figure
\ref{fig:MSE_difference}, we draw an orange cross when MaC outperforms
others and a blue cross otherwise. MaC has superior performance across
the entire range of sample sizes and the number of variables. Similar
to the simulation, the paired t-statistics and Cohen's $d$ for the
real data in the third panel of Table \ref{tbl:sim} show that the
mean and median of MSPE of the MaC are significantly smaller than
those of the other methods. \footnote{{[}}2{]}{Note that the MSPE
of SL for one dataset is extremely large (42.48), which we ignore
in some figures for better visibility (we calculate the paired t-statistic
and Cohen's d after deleting the pair corresponding to this dataset).}

\section{Conclusion\label{sec:conclusion}}

In this paper, we propose a new ensemble learning framework, MaC,
for regression problems. The key feature of MaC being \textit{circular
}\&\textit{ recursive} helps it to communicate among the base machines,
yielding better performance in various scenarios. However, the framework
of making an ensemble algorithm \textit{circular }\&\textit{ recursive}
is not limited to supervised regression learning tasks. Some interesting
extensions would be to apply this same MaC framework to other types
of tasks such as classification and semisupervised learning. In our
theoretical analysis, we derive the risk bound of MaC using a quadratic
risk function. Deriving risk bounds for more general risk functions
and comparing the risks of MaC with individual base machines, super
learner, and boosting could also lead to a deeper understanding of
MaC. We intend to address these extensions in future work. Similar
to boosting methods, MaC has a limitation on the burden of computation,
since the current version of the algorithm of MaC can't be executed
in parallel. It is another important future work to reduce the burden
of computation. 


\section*{Acknowledgments}

Partially supported by NIH Grant 1R21AG074205-01 and NSF Grant DMS-2324489
(Feng) and JSPS KAKENHI (grant number JP19K01582 and JP22H00833) (Liu).

 \bibliographystyle{elsarticle-harv}
\bibliography{ma_mc}

\begin{thebibliography}{23}
\expandafter\ifx\csname natexlab\endcsname\relax\def\natexlab#1{#1}\fi
\providecommand{\url}[1]{\texttt{#1}}
\providecommand{\href}[2]{#2}
\providecommand{\path}[1]{#1}
\providecommand{\DOIprefix}{doi:}
\providecommand{\ArXivprefix}{arXiv:}
\providecommand{\URLprefix}{URL: }
\providecommand{\Pubmedprefix}{pmid:}
\providecommand{\doi}[1]{\href{http://dx.doi.org/#1}{\path{#1}}}
\providecommand{\Pubmed}[1]{\href{pmid:#1}{\path{#1}}}
\providecommand{\bibinfo}[2]{#2}
\ifx\xfnm\relax \def\xfnm[#1]{\unskip,\space#1}\fi
\bibitem[{Breiman(1996)}]{breiman1996bagging}
\bibinfo{author}{Breiman, L.}, \bibinfo{year}{1996}.
\newblock \bibinfo{title}{Bagging predictors}.
\newblock \bibinfo{journal}{Machine learning} \bibinfo{volume}{24},
  \bibinfo{pages}{123--140}.
\bibitem[{Breiman et~al.(1984)Breiman, Friedman, Stone and
  Olshen}]{breiman1984classification}
\bibinfo{author}{Breiman, L.}, \bibinfo{author}{Friedman, J.},
  \bibinfo{author}{Stone, C.J.}, \bibinfo{author}{Olshen, R.A.},
  \bibinfo{year}{1984}.
\newblock \bibinfo{title}{Classification and regression trees}.
\newblock \bibinfo{publisher}{CRC press}.
\bibitem[{Chollet(2017)}]{chollet2017deep}
\bibinfo{author}{Chollet, F.}, \bibinfo{year}{2017}.
\newblock \bibinfo{title}{Deep learning with Python}.
\newblock \bibinfo{publisher}{Simon and Schuster}.
\bibitem[{Cohen(1988)}]{cohen1988statistical}
\bibinfo{author}{Cohen, J.}, \bibinfo{year}{1988}.
\newblock \bibinfo{title}{Statistical power analysis for the behavioral
  sciences}.
\newblock \bibinfo{edition}{2nd} ed., \bibinfo{publisher}{Lawrence Erlbaum
  Associates Inc.}
\bibitem[{Cortes and Vapnik(1995)}]{cortes1995support}
\bibinfo{author}{Cortes, C.}, \bibinfo{author}{Vapnik, V.},
  \bibinfo{year}{1995}.
\newblock \bibinfo{title}{Support-vector networks}.
\newblock \bibinfo{journal}{Machine learning} \bibinfo{volume}{20},
  \bibinfo{pages}{273--297}.
\bibitem[{Dasarathy and Sheela(1979)}]{dasarathy1979composite}
\bibinfo{author}{Dasarathy, B.V.}, \bibinfo{author}{Sheela, B.V.},
  \bibinfo{year}{1979}.
\newblock \bibinfo{title}{A composite classifier system design: Concepts and
  methodology}.
\newblock \bibinfo{journal}{Proceedings of the IEEE} \bibinfo{volume}{67},
  \bibinfo{pages}{708--713}.
\bibitem[{Dong et~al.(2020)Dong, Yu, Cao, Shi and Ma}]{dong2020survey}
\bibinfo{author}{Dong, X.}, \bibinfo{author}{Yu, Z.}, \bibinfo{author}{Cao,
  W.}, \bibinfo{author}{Shi, Y.}, \bibinfo{author}{Ma, Q.},
  \bibinfo{year}{2020}.
\newblock \bibinfo{title}{A survey on ensemble learning}.
\newblock \bibinfo{journal}{Frontiers of Computer Science}
  \bibinfo{volume}{14}, \bibinfo{pages}{241--258}.
\bibitem[{Friedman(2001)}]{friedman2001greedy}
\bibinfo{author}{Friedman, J.H.}, \bibinfo{year}{2001}.
\newblock \bibinfo{title}{{Greedy function approximation: A gradient boosting
  machine.}}
\newblock \bibinfo{journal}{The Annals of Statistics} \bibinfo{volume}{29},
  \bibinfo{pages}{1189 -- 1232}.
\bibitem[{Hastie et~al.(2009)Hastie, Tibshirani and
  Friedman}]{hastie2009elements}
\bibinfo{author}{Hastie, T.}, \bibinfo{author}{Tibshirani, R.},
  \bibinfo{author}{Friedman, J.}, \bibinfo{year}{2009}.
\newblock \bibinfo{title}{The elements of statistical learning: data mining,
  inference, and prediction}.
\newblock \bibinfo{publisher}{Springer Science \& Business Media}.
\bibitem[{Ho(1995)}]{ho1995random}
\bibinfo{author}{Ho, T.K.}, \bibinfo{year}{1995}.
\newblock \bibinfo{title}{Random decision forests}, in:
  \bibinfo{booktitle}{Proceedings of 3rd international conference on document
  analysis and recognition}, \bibinfo{organization}{IEEE}. pp.
  \bibinfo{pages}{278--282}.
\bibitem[{van~der Laan et~al.(2006)van~der Laan, Dudoit and van~der
  Vaart}]{LaanDudoitVaart2006}
\bibinfo{author}{van~der Laan, M.J.}, \bibinfo{author}{Dudoit, S.},
  \bibinfo{author}{van~der Vaart, A.W.}, \bibinfo{year}{2006}.
\newblock \bibinfo{title}{The cross-validated adaptive epsilon-net estimator}.
\newblock \bibinfo{journal}{Statistics \& Decisions} \bibinfo{volume}{24},
  \bibinfo{pages}{373--395}.
\bibitem[{van~der Laan et~al.(2007)van~der Laan, Polley and
  Hubbard}]{vanderLaanPolleyHubbard2007}
\bibinfo{author}{van~der Laan, M.J.}, \bibinfo{author}{Polley, E.C.},
  \bibinfo{author}{Hubbard, A.E.}, \bibinfo{year}{2007}.
\newblock \bibinfo{title}{Super learner}.
\newblock \bibinfo{journal}{Statistical Applications in Genetics and Molecular
  Biology} \bibinfo{volume}{6}, \bibinfo{pages}{1544--6115}.
\bibitem[{Lu and Van~Roy(2017)}]{NIPS2017_49ad23d1}
\bibinfo{author}{Lu, X.}, \bibinfo{author}{Van~Roy, B.}, \bibinfo{year}{2017}.
\newblock \bibinfo{title}{Ensemble sampling}, in: \bibinfo{editor}{Guyon, I.},
  \bibinfo{editor}{Luxburg, U.V.}, \bibinfo{editor}{Bengio, S.},
  \bibinfo{editor}{Wallach, H.}, \bibinfo{editor}{Fergus, R.},
  \bibinfo{editor}{Vishwanathan, S.}, \bibinfo{editor}{Garnett, R.} (Eds.),
  \bibinfo{booktitle}{Advances in Neural Information Processing Systems},
  \bibinfo{publisher}{Curran Associates, Inc.}. pp.
  \bibinfo{pages}{3259--3267}.
\bibitem[{Mendes-Moreira et~al.(2012)Mendes-Moreira, Soares, Jorge and
  Sousa}]{mendes2012ensemble}
\bibinfo{author}{Mendes-Moreira, J.}, \bibinfo{author}{Soares, C.},
  \bibinfo{author}{Jorge, A.M.}, \bibinfo{author}{Sousa, J.F.D.},
  \bibinfo{year}{2012}.
\newblock \bibinfo{title}{Ensemble approaches for regression: A survey}.
\newblock \bibinfo{journal}{Acm computing surveys (csur)} \bibinfo{volume}{45},
  \bibinfo{pages}{1--40}.
\bibitem[{Olson et~al.(2017)Olson, La~Cava, Orzechowski, Urbanowicz and
  Moore}]{Olson2017PMLB}
\bibinfo{author}{Olson, R.S.}, \bibinfo{author}{La~Cava, W.},
  \bibinfo{author}{Orzechowski, P.}, \bibinfo{author}{Urbanowicz, R.J.},
  \bibinfo{author}{Moore, J.H.}, \bibinfo{year}{2017}.
\newblock \bibinfo{title}{Pmlb: a large benchmark suite for machine learning
  evaluation and comparison}.
\newblock \bibinfo{journal}{BioData Mining} \bibinfo{volume}{10},
  \bibinfo{pages}{36}.
\bibitem[{Qi et~al.(2019)Qi, Liu, Wang and Pan}]{NEURIPS2019_3f7bcd0b}
\bibinfo{author}{Qi, Y.}, \bibinfo{author}{Liu, B.}, \bibinfo{author}{Wang,
  Y.}, \bibinfo{author}{Pan, G.}, \bibinfo{year}{2019}.
\newblock \bibinfo{title}{Dynamic ensemble modeling approach to nonstationary
  neural decoding in brain-computer interfaces}, in: \bibinfo{editor}{Wallach,
  H.}, \bibinfo{editor}{Larochelle, H.}, \bibinfo{editor}{Beygelzimer, A.},
  \bibinfo{editor}{d\textquotesingle Alch\'{e}-Buc, F.}, \bibinfo{editor}{Fox,
  E.}, \bibinfo{editor}{Garnett, R.} (Eds.), \bibinfo{booktitle}{Advances in
  Neural Information Processing Systems}, \bibinfo{publisher}{Curran
  Associates, Inc.}. pp. \bibinfo{pages}{6087--6096}.
\bibitem[{Sagi and Rokach(2018)}]{sagi2018ensemble}
\bibinfo{author}{Sagi, O.}, \bibinfo{author}{Rokach, L.}, \bibinfo{year}{2018}.
\newblock \bibinfo{title}{Ensemble learning: A survey}.
\newblock \bibinfo{journal}{Wiley Interdisciplinary Reviews: Data Mining and
  Knowledge Discovery} \bibinfo{volume}{8}, \bibinfo{pages}{e1249}.
\bibitem[{Schapire(1990)}]{Schapire_1990}
\bibinfo{author}{Schapire, R.E.}, \bibinfo{year}{1990}.
\newblock \bibinfo{title}{The strength of weak learnability}.
\newblock \bibinfo{journal}{Machine Learning} \bibinfo{volume}{5},
  \bibinfo{pages}{197--227}.
\bibitem[{Schapire et~al.(1998)Schapire, Freund, Bartlett and
  Lee}]{Schapire1998boosting}
\bibinfo{author}{Schapire, R.E.}, \bibinfo{author}{Freund, Y.},
  \bibinfo{author}{Bartlett, P.}, \bibinfo{author}{Lee, W.S.},
  \bibinfo{year}{1998}.
\newblock \bibinfo{title}{Boosting the margin: A new explanation for the
  effectiveness of voting methods}.
\newblock \bibinfo{journal}{The annals of statistics} \bibinfo{volume}{26},
  \bibinfo{pages}{1651--1686}.
\bibitem[{Sorokina et~al.(2007)Sorokina, Caruana and
  Riedewald}]{sorokina2007additive}
\bibinfo{author}{Sorokina, D.}, \bibinfo{author}{Caruana, R.},
  \bibinfo{author}{Riedewald, M.}, \bibinfo{year}{2007}.
\newblock \bibinfo{title}{Additive groves of regression trees}, in:
  \bibinfo{booktitle}{European Conference on Machine Learning},
  \bibinfo{organization}{Springer}. pp. \bibinfo{pages}{323--334}.
\bibitem[{Sorokina et~al.(2008)Sorokina, Caruana, Riedewald and
  Fink}]{sorokina2008detecting}
\bibinfo{author}{Sorokina, D.}, \bibinfo{author}{Caruana, R.},
  \bibinfo{author}{Riedewald, M.}, \bibinfo{author}{Fink, D.},
  \bibinfo{year}{2008}.
\newblock \bibinfo{title}{Detecting statistical interactions with additive
  groves of trees}, in: \bibinfo{booktitle}{Proceedings of the 25th
  international conference on Machine learning}, pp.
  \bibinfo{pages}{1000--1007}.
\bibitem[{Tian and Feng(2021)}]{tian2021rasea}
\bibinfo{author}{Tian, Y.}, \bibinfo{author}{Feng, Y.}, \bibinfo{year}{2021}.
\newblock \bibinfo{title}{Rase: Random subspace ensemble classification}.
\newblock \bibinfo{journal}{Journal of Machine Learning Research} .
\bibitem[{Yu et~al.(2018)Yu, Luo, Liu, Wong, You, Han and Zhang}]{yu2018semi}
\bibinfo{author}{Yu, Z.}, \bibinfo{author}{Luo, P.}, \bibinfo{author}{Liu, J.},
  \bibinfo{author}{Wong, H.S.}, \bibinfo{author}{You, J.},
  \bibinfo{author}{Han, G.}, \bibinfo{author}{Zhang, J.}, \bibinfo{year}{2018}.
\newblock \bibinfo{title}{Semi-supervised ensemble clustering based on selected
  constraint projection}.
\newblock \bibinfo{journal}{IEEE Transactions on Knowledge and Data
  Engineering} \bibinfo{volume}{30}, \bibinfo{pages}{2394--2407}.

\end{thebibliography}

\pagebreak{}
\begin{table}[htb]
\centering \caption{Results of simulation and real data experiment}
\hspace{0.1in} \label{tbl:sim} \scalebox{0.84}{ %
\begin{tabular}{lrrrrrr}
\toprule 
 & MaC  & SL  & LS-Boost  & DNN  & Tree  & Ridge \tabularnewline
\midrule 
 & \multicolumn{6}{c}{DGP 1}\tabularnewline
\cmidrule(r){1-7} Mean  & 1.92  & 2.21  & 2.11  & 2.48  & 2.53  & 3.36 \tabularnewline
Median  & 1.85  & 2.16  & 2.06  & 2.43  & 2.48  & 3.31 \tabularnewline
paired t  &  & 23.62  & 15.45  & 36.91  & 47.10  & 104.45 \tabularnewline
Cohen's d  &  & 0.75  & 0.49  & 1.17  & 1.49  & 3.30 \tabularnewline
\midrule 
 & \multicolumn{6}{c}{DGP 2}\tabularnewline
\cmidrule(r){1-7} 
 Mean  & 3.05  & 3.21  & 3.27  & 4.02  & 3.35  & 6.25 \tabularnewline
Median  & 2.73  & 2.96  & 2.93  & 3.60  & 3.11  & 5.96 \tabularnewline
paired t  &  & 4.64  & 6.61  & 23.61  & 7.71  & 68.16 \tabularnewline
Cohen's d  &  & 0.15  & 0.21  & 0.75  & 0.24  & 2.16 \tabularnewline
\midrule 
 & \multicolumn{6}{c}{Real data}\tabularnewline
\cmidrule(r){1-7} 
 Mean  & 0.26  & 0.70  & 0.28  & 0.36  & 0.37  & 0.52 \tabularnewline
Median  & 0.20  & 0.27  & 0.23  & 0.32  & 0.32  & 0.61 \tabularnewline
paired t  &  & 6.97  & 12.52  & 33.02  & 15.71  & 26.87 \tabularnewline
Cohen's d  &  & 0.22  & 0.40  & 1.04  & 0.50  & 0.85 \tabularnewline
\bottomrule
\end{tabular}} 
\end{table}

\begin{figure*}
\begin{centering}
\includegraphics[width=0.7\textwidth]{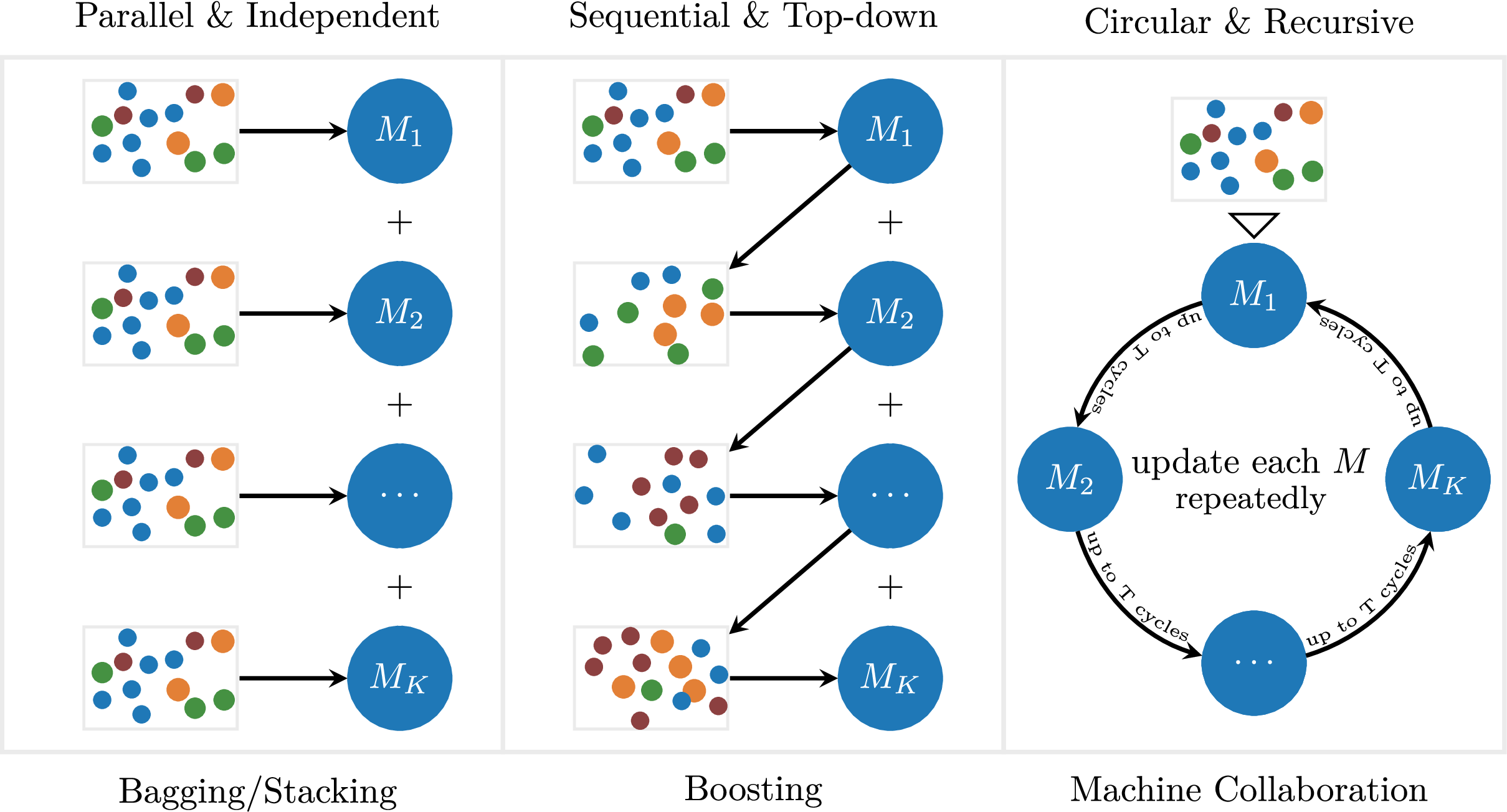} 
\par\end{centering}
\caption{Bagging, boosting, and machine collaboration\label{fig:Bagging,-boosting-and}}
\end{figure*}


\clearpage
\pagebreak
\section*{Appendix}
 The finite sample theoretical result in the main article is an application
of the results of \citet{LaanDudoitVaart2006}. To prove Theorem 1,
we require the following assumptions and proposition.

\paragraph*{Assumptions}
\begin{enumerate}
\item $\underset{\mathfrak{m}\in\mathbb{M}}{\mathrm{sup}}\underset{D\in\mathcal{D}}{\mathrm{sup}}\left|L\left(D,\mathfrak{m}\right)-L\left(D,\mathfrak{m}_{0}\right)\right|\leq C_{1}$.\label{assum:1} 
\item assume\label{assum:2} 
\[
\sup_{\mathfrak{m}\in\mathbb{M}}\frac{VAR_{P_{0}}\left[L(D,\mathfrak{m})-L\left(D,\mathfrak{m}_{0}\right)\right]}{E_{P_{0}}\left[L(D,\mathfrak{m})-L\left(D,\mathfrak{m}_{0}\right)\right]}\leq C_{2}.
\]
\end{enumerate}
\begin{prop} \label{Proposit:1}Under Assumptions 1 and 2,
the following finite sample inequality holds for some constant $C(a)$
\begin{align*}
E_{P_{0}}\delta\left(\hat{\mathcal{M}}\left(P_{n,B_{n}}^{t}\right),\mathcal{\mathfrak{m}}_{0}\right) & \leq(1+a)\min_{k\in\kappa_{n}}\left\{ \left(1+a\right)B_{0}\left(k\right)+C\left(a\right)\frac{1+\log\left(N_{k}\right)}{n\left(1-p\right)}\right\} \\
 & \;\;\;\;\;+C(a)\frac{1+\log\left(J_{n}\right)}{np}
\end{align*}
for any $a>0$. Therein, $N_{k}$ is the searching number of $\mathbb{M}_{k}$
for the MaC algorithm, 
\[
C(a)\equiv4(1+a/2)^{2}\left(\frac{2C_{1}}{3}+\frac{2C_{2}}{a}\right).
\]
\end{prop}

Note that 
\begin{align*}
E_{P_{0}}\delta\left(\hat{\mathcal{M}}\left(P_{n,B_{n}}^{t}\right),\mathcal{\mathfrak{m}}_{0}\right) & =E_{P_{0}}\delta\left(\hat{\mathcal{M}}_{k\left(P_{n}\right)}\left(P_{n,B_{n}}^{t}\right),\mathcal{\mathfrak{m}}_{0}\right)\\
 & =E_{P_{0}}\int\left(\hat{\mathcal{M}}_{k\left(P_{n}\right)}\left(P_{n,B_{n}}^{t}\right)(x)-\mathfrak{m}_{0}(x)\right)^{2}dF_{0}(x).
\end{align*}
Simply applying Theorem 3.1 \citep{LaanDudoitVaart2006} with fixed
$\varepsilon_{k}$, the conclusion of Proposition \ref{Proposit:1}
is straightforward.

\subsection*{Proof of Theorem 1}

We provide a sketch of the proof using the results in \citep{LaanDudoitVaart2006}
here. 
\begin{proof}
To apply Proposition \ref{Proposit:1}, we need only check Assumptions
\ref{assum:1} and \ref{assum:2}. First, Assumption \ref{assum:1}
holds, given 
\begin{align*}
 & \text{\ensuremath{\underset{\mathfrak{m}\in\mathbb{M}}{\mathrm{sup}}\underset{D\in\mathcal{D}}{\mathrm{sup}}\left|L\left(D,\mathfrak{m}\right)-L\left(D,\mathfrak{m}_{0}\right)\right|}}\\
= & \text{\ensuremath{\underset{\mathfrak{m}\in\mathbb{M}}{\mathrm{sup}}\underset{D\in\mathcal{D}}{\mathrm{sup}}\left|\left(Y-\mathfrak{m}\left(x\right)\right)^{2}-\left(Y-\mathfrak{m}_{0}\left(x\right)\right)^{2}\right|}}\\
\leq & \underset{\mathfrak{m}\in\mathbb{M}}{\mathrm{sup}}\underset{D\in\mathcal{D}}{\mathrm{sup}}\left|Y^{2}+\mathfrak{m}^{2}\left(x\right)+2Y^{2}+\mathfrak{m}_{0}^{2}\left(x\right)\right|\\
= & 4C_{0}^{2}=C_{1}.
\end{align*}
For Assumption \ref{assum:2}, note that using some simple algebra,
we have 
\[
\int\left(L(d,\mathfrak{m})-L\left(d,\mathfrak{m}_{0}\right)\right)\mathrm{d}P_{0}(d)=\int\left(\mathfrak{m}(x)-\mathfrak{m}_{0}(x)\right)^{2}\mathrm{d}F_{0}(x).
\]
Then 
\begin{align*}
 & VAR_{P_{0}}\left[L(D,\mathfrak{m})-L\left(D,\mathfrak{m}_{0}\right)\right]\\
\leq & E_{P_{0}}\left[\left(L(D,\mathfrak{m})-L\left(D,\mathfrak{m}_{0}\right)\right)^{2}\right]\\
= & \int\left(\mathfrak{m}^{2}\left(x\right)-\mathfrak{m}_{0}^{2}\left(x\right)-2y\mathfrak{m}\left(x\right)+2y\mathfrak{m}_{0}\left(x\right)\right)^{2}\mathrm{d}P_{0}(d)\\
= & \int\left[\left(\mathfrak{m}\left(x\right)-\mathfrak{m}_{0}\left(x\right)\right)\left(\mathfrak{m}\left(x\right)+\mathfrak{m}_{0}\left(x\right)-2y\right)\right]^{2}\mathrm{d}P_{0}(d)\\
\leq & 16C_{0}^{2}\int\left(\mathfrak{m}\left(x\right)-\mathfrak{m}_{0}\left(x\right)\right)^{2}\mathrm{d}F_{0}(d)\\
= & C_{2}E_{P_{0}}\left[L(D,\mathfrak{m})-L\left(D,\mathfrak{m}_{0}\right)\right].
\end{align*}
Assumption \ref{assum:2} is satisfied. We complete the proof by applying
Proposition \ref{Proposit:1}. 
\end{proof}

\end{document}